\pdfoutput=1
\documentclass[11pt]{article}

\usepackage[]{acl}

\usepackage{times}
\usepackage{latexsym}

\usepackage[T1]{fontenc}
\usepackage[utf8]{inputenc}

\usepackage{microtype}

\usepackage{graphicx}
\usepackage{setspace}
\usepackage{bm}
\usepackage{amsfonts}
\usepackage{amssymb}
\usepackage{booktabs}
\usepackage{multirow}

\usepackage{soul}
\graphicspath{ {./} }
\usepackage{url}

\usepackage{amsmath}

\usepackage{algorithm}
\usepackage{algorithmic}
\usepackage{float}  
\usepackage{lipsum}
\makeatletter
\newenvironment{breakablealgorithm}
{% \begin{breakablealgorithm}
 \begin{center}
  \refstepcounter{algorithm}% New algorithm
  \hrule height.8pt depth0pt \kern2pt% \@fs@pre for \@fs@ruled
  \renewcommand{\caption}[2][\relax]{% Make a new \caption
   {\raggedright\textbf{\ALG@name~\thealgorithm} ##2\par}%
   \ifx\relax##1\relax % #1 is \relax
   \addcontentsline{loa}{algorithm}{\protect\numberline{\thealgorithm}##2}%
   \else % #1 is not \relax
   \addcontentsline{loa}{algorithm}{\protect\numberline{\thealgorithm}##1}%
   \fi
   \kern2pt\hrule\kern2pt
  }
 }{% \end{breakablealgorithm}
  \kern2pt\hrule\relax% \@fs@post for \@fs@ruled
 \end{center}
}
\makeatother

\title{Meta-RTL: Reinforcement-Based Meta-Transfer Learning for Low-Resource Commonsense Reasoning}

\author{
    Yu Fu$^1$\footnotemark[1] \quad 
    Jie He$^1$\thanks{\ \ Equal Contribution.} \quad 
    Yifan Yang$^1$ \quad 
    \textbf{Qun Liu}$^2$ \, and \;
    \textbf{Deyi Xiong}$^1$\thanks{ \, Corresponding  author} \\
    $^1$ College of Intelligence and Computing, Tianjin University, Tianjin, China \\
    $^2$ Huawei Noah’s Ark Lab, Hong Kong, China \\
    \normalsize{\texttt{fuyu\_1998@tju.edu.cn, jieh@ed.ac.uk}} \\
    \normalsize{\texttt{yikfaan.yeung@gmail.com, qun.liu@huawei.com}} \\
    \normalsize{\texttt{dyxiong@tju.edu.cn}} \\
}

\begin{document}

\maketitle

\begin{abstract}
Meta learning has been widely used to exploit rich-resource source tasks to improve the performance of low-resource target tasks. Unfortunately, most  existing meta learning approaches  treat different  source tasks equally, ignoring the relatedness of source tasks to the target task in knowledge transfer.  To mitigate this issue, we propose a reinforcement-based multi-source meta-transfer learning framework (Meta-RTL) for low-resource commonsense reasoning. In this framework, we present a reinforcement-based  approach to dynamically  estimating source task weights that measure the contribution of the corresponding tasks to the target task in the meta-transfer learning. The differences between the general loss of the meta model and task-specific losses of source-specific temporal meta models on sampled target data are fed into the policy network of the reinforcement learning module as rewards. The policy network is built upon LSTMs that  capture long-term dependencies on source task weight estimation across  meta learning iterations.  We evaluate the  proposed Meta-RTL using both BERT and ALBERT as the backbone  of the meta model on three commonsense reasoning benchmark datasets. Experimental results demonstrate that Meta-RTL substantially outperforms strong baselines and previous task selection strategies and achieves larger improvements on extremely low-resource settings. 
\end{abstract}

\section{Introduction}
Commonsense reasoning  is a basic skill of humans to deal with daily situations that involve reasoning about physical and social regularities  \cite{10.1145/2701413}. To endow computers with human-like commonsense reasoning capability  has hence been  one of major goals of artificial intelligence. As commonsense reasoning usually interweaves with many other natural language processing (NLP) tasks (e.g., conversation generation \cite{ijcai2018-643}, machine 
translation \cite{he-etal-2020-box}) and exhibits different forms (e.g., question answering \cite{talmor-etal-2019-commonsenseqa}, co-reference resolution \cite{Sakaguchi_LeBras_Bhagavatula_Choi_2020,long-webber-2022-facilitating,long-etal-2024-multi}), a wide variety of commonsense reasoning datasets have been created recently \cite{Bisk_Zellers_Lebras_Gao_Choi_2020,sap-etal-2019-social}, covering different commonsense reasoning forms and aspects, such as social interaction \cite{sap-etal-2019-social}, laws of nature \cite{Bisk_Zellers_Lebras_Gao_Choi_2020}. 

However, due to the cost of building commonsense reasoning datasets \cite{singh-etal-2021-com2sense,talmor-etal-2019-commonsenseqa}  and the intractability of creating a single unified dataset to cover all commonsense reasoning phenomena, commonsense reasoning  in low-resource settings is vital  for commonsense reasoning tasks with specific forms and limited or no data.  To mitigate this data scarcity issue, a recent strand of research  is 
transfer learning with large pre-trained language models (PLM), where PLMs are further  trained on multiple source datasets and then fine-tuned or directly tested on the target task \cite{DBLP:journals/corr/abs-2103-13009}. Unfortunately,
%节省内容 previous  studies reveal that such multi-source  training of PLMs across multiple datasets does not always guarantee satisfactory results \cite{zhang2020survey,DBLP:journals/corr/abs-1905-05583}. 
as PLMs usually have a large number of parameters and strong memorization power, learning from source-task datasets may force PLMs to memorize useless knowledge of source datasets, causing   negative transfer \cite{yan-etal-2020-multi-source}.

Another promising approach to low-resource NLP is 
meta learning, which allows for better generalization to new tasks \cite{pmlr-v70-finn17a}.  \citet{yan-etal-2020-multi-source} suggests that training a meta-learner for PLMs is  effective to capture transferable knowledge across different tasks. However, this method does not dynamically adjust the weights of source tasks at each iteration  during the meta training for the target task. All source tasks contribute equally to the meta model, which neglects  the distributional heterogeneity in these tasks and different degrees of relatedness of these source tasks to the target task.

\begin{figure*}[!tbp]
\centering
\setlength{\abovecaptionskip}{-0.02cm} 
\setlength{\belowcaptionskip}{-0.3cm} 
%\flushleft
\includegraphics[scale=0.36]{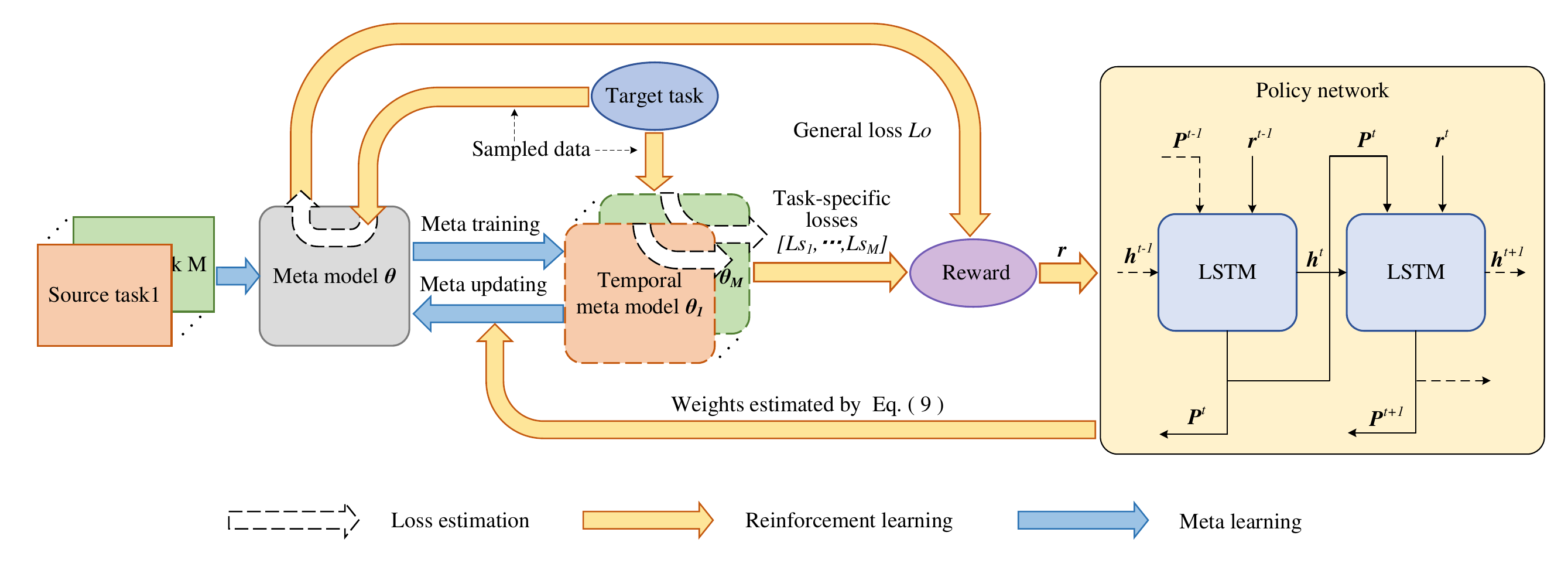}
\caption
{Illustration of Meta-RTL.  An LSTM-based policy network is used to dynamically estimate target-aware weights for source tasks. The estimated weights are explored to update temporal meta models into the meta model in the meta-transfer learning algorithm. The loss differences between the meta model and temporal meta models (source task-specific) on the sampled target task data are fed into the policy network as rewards. } 
\label{figure_main_model}
\end{figure*}

To tackle this issue, we propose a Reinforcement-based Meta-Transfer Learning (Meta-RTL) framework for low-resource commonsense reasoning,   which performs cross-dataset transfer learning  to improve the  adaptability  of the meta model to the target  task. Instead  of fixing source task weights  throughout the entire meta training  process,
We design a policy network, the core component of Meta-RTL, to adaptively estimate a weight for each source task during each meta training iteration. Specifically, as shown in Figure \ref{figure_main_model}, we first randomly sample a batch of tasks from source as source tasks, which are used to train a meta model with a meta-transfer learning algorithm. The meta model is a PLM-based commonsense reasoning model. Once we train a temporal meta model per source task from the meta model, we sample a batch of instances from the target task to evaluate the loss of these temporal meta models on the sampled data. These losses are referred to as task-specific losses. Meanwhile, we also estimate the loss of the meta model on the sampled target data as the general loss.  We use an LSTM-based policy network to predict the weight for each source task. The difference between the task-specific loss and general loss is used as the reward to the policy network. The LSTM nature facilitates the policy network to capture the weight estimation history across meta training iterations. The estimated weights are then incorporated into the meta-transfer learning algorithm to update the meta model. In this way, Meta-RTL is able to learn target-aware source task weights and a target-oriented meta model with weighted knowledge transferred from multiple source tasks.

% We apply Meta-RTL to BERT (Devlin et al.,2019) and ALBERT (Lan et al., 2020) for three sets of commmonsense reasoning tasks: RiddleSense, Creak, Com2sense. Experiments show that Meta-DTL consistently outperforms strong baselines. We also show that Meta-RTL is highly useful for commonsense reasoning tasks when very few training samples of the target task are available. Figure 1 briefly illustrates both meta learning and the proposed Meta-RTL.

To summarize, our contributions are three-fold:
\begin{itemize}
\setlength{\itemsep}{0pt}
\setlength{\parsep}{0pt}
\setlength{\parskip}{0pt}
\item We propose a framework Meta-RTL, which, to  the best of our knowledge,   is the first attempt to explore  reinforcement-based meta-transfer  learning  for low-resource commonsense reasoning.

\item The adaptive reinforcement learning strategy in Meta-RTL facilitates the meta model  to dynamically estimate target-aware weights of source  tasks, which bridges the gap between  the trained meta model  and the target task, enabling a fast convergence on limited target data. 

\item We evaluate  Meta-RTL with BERT-base \cite{devlin-etal-2019-bert} and ALBERT-xxlarge \cite{DBLP:conf/iclr/LanCGGSS20} being used as the backbone of the meta model on three commmonsense reasoning tasks: RiddleSense \cite{lin-etal-2021-riddlesense}, Creak \cite{Onoe-Et-Al-2021-Creak}, Com2sense \cite{singh-etal-2021-com2sense}. Experiments demonstrate that Meta-RTL consistently outperforms strong baselines  by up to 5 points in terms of reasoning  accuracy.
\end{itemize}

\section{Related Work}
The proposed Meta-RTL  is  related to both meta learning and commonsense reasoning, which are reviewed below within the constraint of space. 
\subsection{Meta Learning}
Meta learning, or learning to learn, aims to enhance model generalization and adapt models to new tasks that are not present in training data.
% , is widely used in computer vision \cite{Guo_2020_CVPR,Jamal_2019_CVPR,Liu_2020_CVPR,rajasegaran2020itaml,Wang_2020_CVPR}.
Recent years have gained an increasing attention of meta learning in NLP  \cite{pmlr-v203-he23a}. 
% Gu~\shortcite{gu-etal-2018-meta} use MAML \cite{pmlr-v70-finn17a} to improve low-resource machine translation.  
% Yu~\shortcite{yu-etal-2020-hypernymy} employ meta learning for hypernym detection in low-resource languages. 
% \cite{nooralahzadeh-etal-2020-zero,DBLP:journals/corr/abs-2104-09696} apply meta learning to cross-lingual tasks and examine the unsupervised/supervised capabilities of meta learning in cross-lingual scenarios.
% \cite{yan-etal-2020-multi-source} propose a meta learning method on the monolingual MCQA task to integrate multiple question answering tasks for low-resource QA. Unfortunately, they ignore correlations among different source datasets, and fine-tuning the entire model is not suitable for low-resource tasks as it is prone to overfitting. 
\cite{xiao2020adversarial} propose an adversarial  approach to improving sampling in the meta learning process. Unlike our work, they focus on the same speech recognition task in a multilingual scenario.
\cite{DBLP:journals/corr/abs-2102-09397}  use adapters to perform meta training  on summarization data from different corpora. 
%They use task-specific rules to sample tasks for meta training.  By contrast, our reinforcement-based weight estimation method can be applied to any tasks. 
The most related work to the proposed Meta-RTL is  \cite{yao2021metalearning}. The significant differences from them are two-fold. First,  they focus on different categories under the same task in CV while our interest lies in exploring multiple tasks in commonsense reasoning for the low-resource target task. 
%Second, they  use task loss as input to the policy network to learn a general representation while we employ the differences in task losses as rewards. 
Second,  they simply utilize  MLP to estimate  weights  at each step. In contrast, we   use LSTM to encode the long-term information  across  training iterations  to calculate adaptive weights. 
To sum up,  previous works either mechanically use a fixed task sampling  strategy or just take into account the variability of different original tasks. Substantially different from them, we propose a reinforcement-based  strategy to  adaptively estimate target-aware weights for source tasks in the meta-transfer learning in order to enable weighted knowledge transfer. 

\subsection{Commonsense Reasoning  and Datasets}

A wide range of  commonsense reasoning datasets have been proposed  recently. \cite{gordon-etal-2012-semeval} create COPA for causal inference while \cite{rahman-ng-2012-resolving} present Winogrand Scheme Challenge (WSC), a dataset testing commonsense reasoning in the form of anaphora resolution. Since the size of these datasets is usually small, effective training cannot be obtained until the recent emergence of pre-training methods  \cite{he-etal-2019-hybrid}. On the other hand,  large commonsense reasoning datasets have been also curated  \cite{Sakaguchi_LeBras_Bhagavatula_Choi_2020,sap-etal-2019-social,huang-etal-2019-cosmos,long-etal-2020-ted,long-etal-2020-shallow}, which facilitate the training of  neural commonsense reasoning models. 
A popular trend to deal with these datasets is using graph neural networks for reasoning with external KGs \cite{feng-etal-2020-scalable,he-etal-2022-evaluating,he-etal-2023-buca,he2025evaluatingimprovinggraphtext,he2025mintqamultihopquestionanswering,long2024leveraginghierarchicalprototypesverbalizer}, and fine-tuning unified text-to-text QA models \cite{khashabi-etal-2020-unifiedqa}. 
Apart from ConceptNet,  Wikipedia and Wiktionary are also used as additional knowledge sources for commonsense reasoning \cite{xu-etal-2021-fusing}. 
% A few recent methods also aim to generate relevant triples via language generation models so that the context graph is more beneficial for reasoning (Wang et al., 2020; Yan et al., 2020).
RAINBOW \cite{DBLP:journals/corr/abs-2103-13009}, which uses multi-task learning 
to provide a pre-trained commonsense reasoning model on top of various large-scale commonsense reasoning datasets, is related  to our work. However,  RAINBOW  only performs multi-task learning, which does not aim at knowledge transfer to a low-resource target task.

\section{Meta-RTL}
\label{sec3}
The proposed reinforcement-based meta-transfer learning framework for low-resource commonsense reasoning is illustrated in Figure \ref{figure_main_model}. It consists of three essential components: a PLM-based commonsense reasoning model, a meta-transfer learning algorithm that trains the PLM-based commonsense reasoning model and a  reinforcement-based target-aware weight estimation strategy that is equipped to the meta-transfer learning algorithm for estimating source task weights.

\subsection{PLM-Based Commonsense Reasoning Model}
\label{riddlesense_reason}
Commonsense reasoning tasks are usually in the form of multiple-choice question answering. We hence choose  a masked language model as the commonsense reasoning backbone  to predict answers. However, as different commonsense reasoning datasets differ in  the  number  of  candidate answers  (e.g., 2 candidate answers per question  in Com2sense vs 5 in CommonseseQA), a PLM classifier with a fixed number of classes is not a good fit for this scenario. To tackle this issue, partially inspired by \cite{sap-etal-2019-social},  For each  candidate answer, we concatenate it with context, question into $[ \rm{CLS}] \langle \rm{context} \rangle \langle \rm{question}\rangle [ \rm{SEP}]\langle \rm answer_{i} \rangle[ \rm{SEP}] $, where [CLS] is a special token for aggregating information while [SEP] is a separator. We stack a multilayer perceptron over the backbone to compute a score $ \hat y_i$ for $\rm answer_i$, with the hidden state $ \bm{h}_{\rm{CLS}}\in \mathbb{R}^H$: 
\begin{equation}
     \hat y_i=\bm{W}_2 \tanh(\bm{W}_1\bm{h}_{\text{CLS}}+\bm{b}_1)
\end{equation} 
 where $\bm{W}_1 \in \mathbb{R}^{H \times H},\bm{b}_1\in \mathbb{R}^H, \bm{W}_2 \in \mathbb{R}^{1\times H}$ are learnable parameters and $H$ is the dimensionality.
 
Finally, we estimate the probability distribution over candidate answers using a softmax layer: 
\begin{equation}
    \bm{Y}=\text{softmax}([\hat y_1,...,\hat y_N])
\end{equation}where $N$ is the number of candidate answers. 
The final answer predicted by the model corresponds to the context-answer pair with the highest probability.

This PLM-based commonsense reasoning model is used as the meta model that is trained in the meta-transfer learning algorithm described in the next subsection.

\subsection{Meta-Transfer Learning Algorithm}

The training procedure for the meta model is illustrated in Algorithm \ref{algorithm1}, which is composed of two parts: meta learning over multiple source tasks and transfer learning to the target task.

\subsubsection{Meta Learning over Multiple Source Tasks} 
\label{meta-learning}
%  (i.e., iteration in the outer loop in lines 1-17)
The meta learning procedure is presented in lines 1-19 in Algorithm \ref{algorithm1}. For each meta training iteration, we use $M$ source datasets. For each source dataset $s_i$, we randomly sample instances from it to construct a task $\mathcal{T}_{s_i}$ for meta training, which is then randomly split into two parts: support set $\mathcal{T}_{s_i}^{\text{sup}}$ and query set $\mathcal{T}_{s_i}^{\text{qry}}$, which do not overlap each other. All source tasks are denoted as $\mathcal{T}_s = \{\mathcal{T}_{s_1},\mathcal{T}_{s_2},...,\mathcal{T}_{s_M} \}$. The learning rates for the inner and outer loop in the algorithm are different: $ \alpha$ denotes the learning rate for the inner loop, while $ \beta$ for the outer loop.

The inner loop (lines 4-8) aims to learn source information from different source datasets. For each source task $\mathcal{T}_{s_i}$, the task-specific parameters $ \bm{\theta}_{\mathcal{T}_{s_i}}$ (i.e., the temporal meta model as illustrated in Figure \ref{figure_main_model}) are updated as follows: 
\begin{equation}
    \bm{\theta}_{\mathcal{T}_{s_i}} =\bm{\theta}-\alpha\nabla_{\bm{\theta}} \mathcal{L}_{\mathcal{T}_{s_i}^{\text{sup}}}(f(\bm{\theta}))
    \label{meta_support}
\end{equation}
where the loss function $ \mathcal{L}_{\mathcal{T}_{s_i}^{\text{sup}}}(f(\bm{\theta}))$ is calculated by fine-tuning the meta model parameters $\bm{\theta}$ on the support set $\mathcal{T}_{s_i}^{\text{sup}}$.

In the outer loop, $ \mathcal{L}_{\mathcal{T}_{s_i}^{\text{qry}}}(f(\bm{\theta}_{\mathcal{T}_{s_i}}))$ is calculated with respect to $ \bm{\theta}_{\mathcal{T}_{s_i}}$, to update the meta model on the corresponding query set $\mathcal{T}_{s_i}^{\text{qry}}$.

It is worth noting that $ f(\bm{\theta}_{\mathcal{T}_{s_i}})$ is an implicit function of $ \bm{\theta}$. As the second-order Hessian gradient matrix requires  expensive computation, we employ the Reptile \cite{DBLP:journals/corr/abs-1803-02999} algorithm, which ignores second-order derivatives and uses  the difference between $\bm{\theta}$ and $\bm{\bm{\theta}}_{\mathcal{T}_{s_i}}$ as the gradient to update the meta model:
\begin{equation}
   \bm{\theta}=\bm{\theta} + \beta\frac{1}{M}\sum_{i=1}^{M}(\bm{\theta}_{\mathcal{T}_{s_i}} - \bm{\theta})
   \label{meta_query}
\end{equation}

We keep running the meta learning procedure until the  meta model converges. By meta learning, we can learn a general meta space, from which we induce meta representations, mapped by the meta model from the source datasets to the meta space. As the meta model is trained across multiple source tasks, the learned meta representations are of generalization capability. 
 \begin{breakablealgorithm}
    \caption{Meta-Transfer Learning Algorithm}
    \setstretch{1}

    \begin{algorithmic}[1]
    \begin{footnotesize} %%调整算法字体⼤⼩
        \textbf{Inputs}: \\
	    Task distribution over source datasets $ p(\mathcal{T}_s)$; \\
		Data distribution of the target dataset $ p(\mathcal{T}_t)$;\\
	    \textbf{Parameters}: \\
	    Parameters $\bm{\theta}$ of the pretrained metal model; \\
	    Parameters $\bm{\phi}$ of the policy network; \\
	    Inner-loop learning rate $\alpha$, outer-loop learning rate $\beta$, transfer learning rate $\gamma$;
        \WHILE{not done}  % while的开始 
		\STATE  Sample source tasks $ \mathcal{T}_{s_j}^{\text{(}i\text{)}} \sim p(\mathcal{T}_{s_j})$ to obtain $ \{\mathcal{T}_{s_j}^{\text{(}i\text{)}}\}_{j=1}^M$ for the current iteration ($i$)
		\STATE Sample data $\mathcal{D}^{\text{(}i\text{)}}_t \sim p(\mathcal{T}_t)$ from the target dataset for the current iteration ($i$) and compute the general loss $\mathcal{L}_{o}$ of the meta model on the sampled target data $\mathcal{D}^{\text{(}i\text{)}}_t$ 
		%$d^{\text{(}i\text{)}}$
		\FOR {all $ \{\mathcal{T}_{s_j}^{\text{(}i\text{)}}\}_{j=1}^{M}$}
		\STATE Fine-tune the meta model on the support set $ \mathcal{T}_{s_j}^{\text{sup}}$ in $ \mathcal{T}_{s_j}^{\text{(}i\text{)}}$ to update parameters:
		\STATE \quad $ \bm{\theta}_{\mathcal{T}_{s_j}}=\bm{\theta}-\alpha \nabla_{\bm{\theta}} 
	    L_{\mathcal{T}_{s_j}^{\text{sup}}}(f(\bm{\theta}))$
		\STATE Compute the task-specific loss $\mathcal{L}_{s_j}$ using $\mathcal{D}^{\text{(}i\text{)}}_t$ for $\bm{\theta}_{\mathcal{T}_{s_j}}$
		\ENDFOR
		\STATE  Get sample probabilities using $ f(\bm{\phi})$: 
		\STATE \quad $ \bm{P}=(P_{1},P_{2},\dots,P_{M})$
		\STATE Get sampled $N$ trajectories according to Eq. (\ref{rl_prob}):
		\STATE$\quad \bm{\tau} = (\tau^1, \tau^2, \dots, \tau^N)$
		\STATE  Compute source task weights according to Eq.  (\ref{get_rate}):
		\STATE \quad $\bm{C} = (C_1, C_2, \dots, C_M)$ 
		\STATE Update   $ \bm{\theta}=\bm{\theta} + \beta \sum_{j=1}^{M} C_{i} \cdot (\bm{\theta}_{\mathcal{T}_{s_j}} - \bm{\theta})$
		\STATE Compute rewards according to Eq. (\ref{get_reward}):
	    \STATE \quad $\bm{r} = (r_1, r_2, \dots, r_M)$
		\STATE Update $\bm{\phi}$ according to Eq. (\ref{rl_estimate})
		\ENDWHILE
		\\
	    \STATE Sample mini-batches dataset $\{o^{k}\}_{k=1}^{B} \sim p(\mathcal{T}_t)$ from the target dataset
		\FOR {all \{ $o^k\}_{k=1}^{B}$}
		\STATE Calculate gradients on the meta model $ \nabla_{\bm{\theta}} L_{o^k}(f(\bm{\theta}))$
		\STATE Update $\bm{\theta} = \bm{\theta} - \gamma \nabla_{\bm{\theta}} \mathcal{L}_{o^k}(f(\bm{\theta}))$
		\ENDFOR
    \end{footnotesize}
    \end{algorithmic}
    \label{algorithm1}
\end{breakablealgorithm}

 %\end{algorithm}

\subsubsection{Transfer Learning to the Target Task} 
The transfer procedure is presented in lines 20-24 in Algorithm \ref{algorithm1}. After performing meta learning, the transfer module will be applied upon the meta model to bridge the gap between the learned meta representations and the data distribution  space of the target dataset. We use the training data of the target task to fine-tune the meta model trained in the meta learning procedure.

%\vspace{13pt}
 %\begin{algorithm}

\subsection{Reinforcement-Based Target-Aware Weight Estimation Strategy}

% 节省空间 \subsubsection{Motivation} 
% For meta learning in NLP, it is crucial to properly define source tasks for meta training. Previous approaches usually use a static method to choose source datasets and then construct source tasks, e.g., according to  transfer accuracy \cite{yan-etal-2020-multi-source}. Tasks from all chosen datasets are then equally explored during each meta training iteration. This is not a good strategy since different source datasets might have different data distributions from the target dataset. A single undesirable task in the chosen source tasks might optimize the model towards a wrong direction. We therefore propose a reinforced strategy to dynamically estimate weights for source tasks in meta training. Through the proposed approach, we can deal with the distributional heterogeneity issue between source tasks and the target task and enable target-aware weighted knowledge transfer from source tasks to the target task. 

% 节省空间\subsubsection{Target-Aware Task Estimation Approach} 

For each meta training iteration, we calculate a general loss $ \mathcal{L}_{o}$ on the meta model $ f(\bm{\theta})$ using sampled data from the target dataset (line 3). After optimizing $ f(\bm{\theta})$ with $\mathcal{T}_{s_i}^{\text{sup}}$  according to Eq. (\ref{meta_support}), we obtain a task-specific model $f(\bm{\theta}_{\mathcal{T}_{s_{i}}})$ for each source task together with a task-specific loss $ \mathcal{L}_{s_{j}}$ on the same sampled data as $\mathcal{L}_o$ (line 7).

To dynamically weight source tasks, we use the difference between the general loss $\mathcal{L}_o$ and task-specific loss $\mathcal{L}_{s_j}$ as a guiding signal. Such a difference can measure how good the meta model is for the target dataset after being tuned by the corresponding source task. In the traditional meta training as formulated in Eq. (\ref{meta_query}), all task-specific models are treated equally. Inspired by \cite{xiao2020adversarial}, we use an LSTM-based network together with an FFN and attention layer to capture the long-term dependencies on historical weight estimation across meta training iterations. Since we do not have any annotated data to train the LSTM-based network, we use REINFORCE \cite{DBLP:journals/ml/Williams92}, a policy gradient algorithm, for our proposed reinforcement-based source task weight estimation and use the guiding signal as the reward.

Let $ f_{\bm{\phi}}(\cdot)$ denote the LSTM-based network trained by reinforcement learning, $ \bm{\phi}$ be parameters to be tuned and $r_j$ as the difference for the $j$-th source task, computed as follows:
\begin{equation}
 r_{j} = \mathcal{L}_o - \mathcal{L}_{s_j}
\label{get_reward}   
\end{equation}
For REINFORCE training,  at each meta training iteration $t$, we feed $ \bm{r}^{t-1} = (r^{t-1}_{1}, r_{2}^{t-1}, \dots, r_{M}^{t-1})$ into the policy network together with the probabilities $\bm{P}^{t-1}=(P_{1}^{t-1}, P_{2}^{t-1}, \dots, P_{M}^{t-1})$, which are estimated in the previous step for the policy network. We then obtain a new updated probability distribution over source tasks denoted as $ \bm{P}^t = (P_{1}^{t}, P_{2}^{t}, \dots, P_{M}^{t})$ from the output of the policy network. In the meantime, we have updated $\bm{r}^t = (r_{1}^{t}, r_{2}^{t}, \dots, r_{M}^{t})$ accordingly. We treat the estimation of the weights for the source tasks as a contextual bandit problem as in \cite{DBLP:conf/emnlp/DongSCHC18}. Formally, for the source tasks $\{ \mathcal{T}_{s_1}, \mathcal{T}_{s_2}, \dots, \mathcal{T}_{s_M}\}$, we sample $K$ tasks $\tau=\{\mathcal{T}_{\tau_1},\mathcal{T}_{\tau_2}, \dots, \mathcal{T}_{\tau_K} \}$ as a trajectory to compute rewards, where $\tau_k \in \{s_1, s_2, \dots, s_M\}$ and $K$ is an integer hyper-parameter. The gradients to update the policy network can be calculated as:
% $\{ \mathcal{T}_{s_m}\}_{m=1}^{M}$
%

\begin{equation}
     \nabla_{\bm{\phi}}J(\bm{\phi}) \approx  \frac{1}{N} \sum_{n=1}^{N} \nabla_{\bm{\phi}}(R(\tau^{n}) - \Tilde{r}))\log f_{\bm{\phi}}(\tau^{n})
\label{rl_estimate}
 \end{equation}

 where $\Tilde{r}$ is the baseline value to reduce the variance in the REINFORCE algorithm, $\tau^n$ is the  $ n$-th sampled trajectory in the total $N$ sampled trajectory, $R(\tau^n) = \sum_{k=1}^{K} r_{\tau^n_k}^{t}$ denotes the rewards of the trajectory. 
 
As shown in Eq. (\ref{rl_estimate}), we use the sampled trajectories to collect rewards and gradients to update the policy network. This procedure might quickly converge to a local minimum and the policy would become a deterministic policy. To avoid this problem, we incorporate the $\rm \epsilon$-greedy technique into the sampling process and entropy regularization into the gradient calculation.

\begin{table*}
\centering
\small
\setlength{\abovecaptionskip}{-0.05cm} 

 \setlength{\belowcaptionskip}{-16pt} 

\begin{tabular}{ccccccc}
\hline
                          & \multicolumn{2}{c}{Com2sense}                                         & \multicolumn{2}{c}{Creak}                                     & \multicolumn{2}{c}{RiddleSense}                               \\
\multirow{-2}{*}{Methods} & unsupervised                          & supervised                                  & unsupervised                          & supervised                          & unsupervised                          & supervised                          \\ \hline
Random&50.00&50.00&50.00&50.00&20.00&20.00\\
Target Fine-tuning (BERT)                      & -                         & 54.22                                 & -                         & 69.80                          & -                         & 56.22                         \\
Reptile (BERT)             & 54.48                         & 57.03                                 & 55.43                         & 67.47                         & 35.26                         & 56.42                         \\
Task Comb. (BERT)&55.24&58.31&57.11&68.49&36.24&54.06\\
Temp. Reptile (BERT)&55.75&58.44&56.75&68.56&37.32&57.49\\
\textbf{Meta-RTL (BERT)}            & \textbf{56.78}                & {\color[HTML]{333333} \textbf{59.08}} & \textbf{58.57}                & \textbf{71.48}                & \textbf{38.10}                & \textbf{58.86}                \\ \hline
Random&50.00&50.00&50.00&50.00&20.00&20.00\\
Target Fine-tuning (ALBERT)                    & -                         & {\color[HTML]{333333} 57.03}          & -                        & 81.55                         & -                         & 71.40                         \\
Reptile (ALBERT)           & 61.64                         & {\color[HTML]{333333} 69.95}          & 69.75                         & 79.87                         & 48.09                         & 70.71                         \\
Task Comb. (ALBERT)&65.72&68.80&68.13&77.46&51.32&72.67\\
Temp. Reptile (ALBERT)&65.47&71.48&68.34&79.64&48.77&70.62\\

\textbf{Meta-RTL (ALBERT)}        & \textbf{66.62}                & {\color[HTML]{333333} \textbf{72.38}} & \textbf{71.26}                & \textbf{82.06}                & \textbf{53.48}                & \textbf{74.44}                \\ \hline
\multicolumn{1}{l}{}      & \multicolumn{1}{l}{\textbf{}} & \multicolumn{1}{l}{\textbf{}}         & \multicolumn{1}{l}{\textbf{}} & \multicolumn{1}{l}{\textbf{}} & \multicolumn{1}{l}{\textbf{}} & \multicolumn{1}{l}{\textbf{}}
\end{tabular}
 \caption{Accuracy results on the 3 datasets using BERT and ALBERT as backbone models. ``-'' indicates no such combination.}
\label{table_main}
\end{table*}

The $\epsilon$-greedy technique regards the sampling process as a progressive process where previous sampling affects the probability of succeeding sampling. The log of the trajectory  which is required in Eq. (\ref{rl_estimate}) is hence computed as follows:
% \begin{equation}
% \begin{split}
%     \begin{small}
%     \log f_{\bm{\phi}}(\tau^{n}) = \log \big(\prod_{k=1}^{K}(\frac{\epsilon}{M - k + 1}    \end{small}
%  \\     \begin{small}
% +\frac{(1 - \epsilon) * P_{\tau_{k}^{n}}^{t}}{ 1 - \sum_{z=1}^{k-1} P_{\tau_{z}^{n}}^{t}})\big)
%     \label{rl_prob}
%     \end{small}
% \end{split}
% \end{equation}

\begin{equation}
    \log f_{\bm{\phi}}(\tau^{n})  \!= \!\log \!\big(\!\prod_{k=1}^{K}(\frac{\epsilon}{\!M \!- \!k \!+ \!1} + \frac{(1 \!-\!  \epsilon) * P_{\tau_{k}^{n}}^{t}}{ 1 \!- \!\sum_{z=1}^{k-1} \!P_{\tau_{z}^{n}}^{t}})\big)
    \label{rl_prob}
\end{equation}

By setting  $\epsilon$, we can control source task probability estimation.  Large $\epsilon$ indicates a high probability towards random sampling, which leads to a high exploration rate. 
% The $\epsilon$-greedy technique can be seen as the trade-off between exploration and exploitation.

For entropy regularization, we use the probability distribution $\bm{P}^t$ estimated by the policy network to calculate the entropy and combine it into the policy network updating as:
 \begin{equation}
      \nabla_{\bm{\phi}}J(\bm{\phi}) = \nabla_{\bm{\phi}}J(\bm{\phi})  + \rho \nabla _{\bm{\phi}} \sum_{m=1}^{M} (- P^{t}_{m} \log P^{t}_{m})
\label{rl_estimate_final}
 \end{equation}
 where $\rho$ is to control the rate of the entropy in the updating gradient.

We average over the multiple sampled trajectories to estimate the weights of source tasks $\bm{C} = (C_1, C_2, \dots, C_M)$ which can be calculated as:

\begin{equation}
    \bm{C} = \frac{1}{NK}\sum_{n=1}^{N} \sum_{k=1}^{K}(C_{\tau^{n}_{k}}+1)
    \label{get_rate}
\end{equation}
where $\tau^{n}_{k}$ denotes the $k$-th chosen task in the $n$-th trajectory  obtained from Eq. (\ref{rl_prob}).

We finally integrate the estimated weights into the meta training stage to bridge the gap between the learned meta representations and the target dataset distribution as follows:

\begin{equation}
   \bm{\theta}=\bm{\theta} + \beta \sum_{i=1}^{M} C_{i} \cdot (\bm{\theta}_{\mathcal{T}_{s_i}} - \bm{\theta})
   \label{meta_query_rl}
\end{equation}

In summary, we train the meta learning module and reinforcement-based weight estimation module together. For the meta training, we obtain weights from the reinforcement-based estimation model. We then follow the meta learning procedure described in Section \ref{meta-learning} and calculate gradients according to Eq. (\ref{meta_query_rl}).  For the reinforcement-based estimation module, at timestep $t$, the module collects probabilities $\bm{P}^{t-1}$ and rewards $\bm{r}^{t-1}$  in the previous timestep $t - 1$ as the inputs, and outputs the current task probabilities $\bm{P}^{t}$. Using the current rewards $\bm{r}^{s}$ and $\bm{P}^{t}$, we update the policy network according to Eq. (\ref{rl_estimate}), Eq. (\ref{rl_prob}) and Eq. (\ref{rl_estimate_final}).

\section{Experiments}
We conducted experiments  using 5 commonsense reasoning benchmark datasets  and examined the effectiveness of the proposed model on 3 latest datasets (i.e., Com2sense, Creak and RiddleSense). For each dataset to be evaluated, we chose this dataset as the target dataset while the other 4 datasets as the source datasets.  The details for datasets and experimental settings are provided in  Appendix \ref{dataset_appendix}  and \ref{setting_appendix}.

We compared our proposed  Meta-RTL against  the following 4  baselines:
\begin{itemize}
\setlength{\itemsep}{0pt}
\setlength{\parsep}{0pt}
\setlength{\parskip}{0pt}
    \item \textbf{Target Fine-tuning} that uses the training  data in the target dataset to fine-tune  the backbone model (BERT-base and ALBERT-xxlarge).
    \item \textbf{Reptile} \cite{DBLP:journals/corr/abs-1803-02999} that uses  the Reptile algorithm to train the meta model without any changes.
    \item \textbf{Task Combination} that combines all the source datasets together and uses the merged dataset to train the backbone model. 
    \item \textbf{Temperature-based Reptile} \cite{tarunesh-etal-2021-meta} that estimates the sample probability as  $P_{m} = d_{i}^{1/\omega} / (\sum_{m=1}^{M} d_{m}^{1 / \omega})$ where $d_i$ is the size of the $i-$th source dataset and $\omega$ is the temperature hyperparameter.

\end{itemize}
% \paragraph{Single-task} we use the train data in the target dataset to fine tune on the backbone model (BERT and ALBERT). Through some datasets have the results in its papers, we reproducted the results for a fair comparsion.

% \paragraph{Meta-train} we use the Reptile algorithm to train the meta model without any modified. In the table \ref{table_main}, we use ``Reptile'' denote this baseline. For own method is based on the Reptile algorithm and use the reinforce-based module to learn a weights. 

% Note that, the Target Fine-tuning result may be different from the results in the original paper. For example, we get 54.22 for Com2sense dataset on BERT, but the result report on the paper is 57.07, But for Riddlesense datset, we get 71.40 on ALBERT, while the result in the paper is 66.99. So we use own results as the final baseline and use the same fine tune procedure to get the supervised results on the Reptile and own proposed Meta-RTL method. For evaluation, we use the prediction accuracy as the evaluation metric and get the supervised result on the dev-data from the datasets.
% For a fair comparison, we reproduced all results for Single-task based on our own implementation, which may be different from results reported in previous work. 
As test sets are not publicly available, we report  accuracy results on development sets.  On each target dataset, we reported results under two settings: supervised and unsupervised. The former used the corresponding target dataset to fine-tune the trained commonsense reasoning model  while the latter did not. The complexity analysis of our method against the baselines is provided in Appendix \ref{complexity_appendix}.

\subsection{Main Results}
Main results are displayed in Table \ref{table_main}. From the table, we observe that:
\begin{itemize}
\setlength{\itemsep}{0pt}
\setlength{\parsep}{0pt}
\setlength{\parskip}{0pt}
\item Our proposed Meta-RTL  significantly outperforms the four baselines under both supervised and unsupervised settings across all three datasets. On Com2sense,  Meta-RTL  under the unsupervised setting is even much better than the target fine-tuning method under the supervised setting by up to 9.59 points with ALBERT (66.62 vs. 57.03).
\item Heuristic methods including Task Comb. and Temp. Reptile cannot  always boost the performance, e.g.,  results of  both BERT (69.80 vs. 68.49) and ALBERT (81.55 vs. 77.46)  on the Creak dataset. The Temp. Reptile is  not always better than Reptile (e.g.,  70.62 vs. 70.71 with ALBERT on the Riddlesense dataset).

\item Meta-RTL is able to steadily achieve substantial improvements over the Four strong baselines no matter what backbone model is used for commonsense reasoning. Although ALBERT is much better than BERT for commonsense reasoning on all three datasets, the improvements of Meta-RTL over Reptile on ALBERT are comparable to those on BERT (e.g., 2.43 vs. 2.05 on Com2sense and 3.73 vs. 2.44 on RiddleSense), indicating that Meta-RTL is  robust to different PLM-based backbone models  to some extent and may  benefit from the size of the model.
\item On the three target datasets, the smaller the target dataset is, the larger  the improvement over target  fine-tuning under the supervised setting is achieved by Meta-RTL (i.e, 1.68 on Creak, 2.64 on RiddleeSense and 4.86 on Com2sense). This suggests that Meta-RTL is beneficial to low-resource commonsense reasoning. 

\end{itemize}
\begin{table}[]
 \setlength{\belowcaptionskip}{-16pt} 
\small 
    \centering
    \begin{tabular}{lcc}
    \bottomrule
         Method&Unsupervised&Supervised  \\
         \hline 
        %  Reptile&\textbf{54.48}&\textbf{57.03}\\
        %  FOMAML&53.71&56.27\\
        %  \hline
        %  Temp. Reptile&\textbf{55.75}&\textbf{58.44}\\
        %  % Temp. FOMAML&53.07&55.63\\
        %  Temp. FOMAML & 52.43 & 56.14 \\
        %  \hline
        %  Meta-RTL (Reptile)&\textbf{56.78}&\textbf{59.08}\\
        %  Meta-RTL (FOMAML)&54.73&57.29\\
         FOMAML&53.71&56.27\\
         Reptile&\textbf{54.48}&\textbf{57.03}\\
         \hline
         % Temp. FOMAML&53.07&55.63\\
         Temp. FOMAML & 52.43 & 56.14 \\
         Temp. Reptile&\textbf{55.75}&\textbf{58.44}\\
         \hline
         Meta-RTL (FOMAML)&54.73&57.29\\
         Meta-RTL (Reptile)&\textbf{56.78}&\textbf{59.08}\\
         \toprule
    \end{tabular}
    \caption{Comparison of different meta learning methods on Com2sense.}
    \label{tab:fomaml}
\end{table}
% \vspace{-5pt}
\subsection{Evaluation with Different  Meta-Learning Algorithms}
We further conducted experiments with  two different widely-used  meta learning algorithms to validate the effectiveness  of our proposed method.

Results with FOMAML \cite{pmlr-v70-finn17a} and Reptile \cite{DBLP:journals/corr/abs-1803-02999} are shown in Table \ref{tab:fomaml}. Our proposed method is able to improve both meta learning methods. However, the Temperature-based method fails to improve  FOMAML  (see supervised/unsupervised results of Temp. FOMAML vs. FOMAML in Table \ref{tab:fomaml}), which demonstrates that our proposed method is more flexible and can be dynamically adapted during the learning procedure. 

As shown in Table \ref{tab:fomaml}, Meta-RTL (Reptile) is better than   Meta-RTL (FOMAML)  under both supervised and unsupervised settings. We conjecture that this could be due  to our reward calculation method. Reptile directly uses the model parameters to calculate the update gradient which is more closely related to the general loss $\mathcal{L}_o $ and the task-specific loss $\mathcal{L}_{s_j}$ than the query loss gradient used in  FOMAML. We therefore use  Reptile as the meta learning  algorithm  in subsequent experiments.

\subsection{Ablation Study on the Weight Estimation Approach}
We further compared with several other methods to examine the effectiveness of the proposed reinforcement-based weight estimation approach. 

Results are shown in Table \ref{re-ablation} (using Com2sense as the target dataset). ``TL'' indicates pure transfer learning from one or multiple source datasets to the target dataset. Specifically, we pretrain the backbone model on specified source datasets and then fine-tune it on the target dataset. As we can see, pure transfer learning is not always able to improve performance over the direct fine-tuning on the target dataset (i.e., Target Fine-tuning in Table \ref{re-ablation}). Furthermore, simply putting all  source datasets together for transfer learning (denoted  as Task Comb.)), despite achieving improvements over the target fine-tuning, is still inferior  to our proposed method. And the Task Comb.  cannot  always perform well on all  datasets, as shown in Table \ref{table_main}.
% is worse than transfer learning from a single source dataset (i.e., Creak), which suggests that simple combination of different source datasets for transfer learning is not optimal. 
\begin{table}[]
\centering
\setlength{\belowcaptionskip}{-16pt}

\small
\begin{tabular}{ccc}
\hline
Method              & unsupervised  & supervised   \\
\hline
TL (C)              & 49.62 & 53.32 \\
TL (R)              & 49.74 & 54.60  \\
TL (W)              & 50.00 & 56.65 \\
TL (Cr)             & 51.02 & 57.03 \\
% TL (C + R + W + Cr) & 52.94 & 56.52 \\
Task Comb.           & 55.24 & 58.31 \\
Reptile             & 54.48	& 57.03 \\
Random              & 53.45 & 57.16 \\
Greedy              & 54.86 & 57.29 \\
Our                 & \textbf{56.78} & \textbf{59.08} \\
\hline
\end{tabular}
\caption{Ablation study results on Com2sense  with BERT as the backbone model.  ``C'': CommonseseQA. ``R'': RiddleSense. ``W'': Winogrande. ``Cr'': Creak. TL (*)  denotes transfer learning from  the corresponding dataset  to the target task.}
\label{re-ablation}
\end{table} 

\begin{table*}[!tbp]

 \centering
 \small
% \begin{tabular}{c|ccccc}
% \hline
% {Percentage} & { Target Fine-tuning} &{ Reptile} & { Meta-RTL} & {  $\rm \delta$ (Target Fine-tuning)} & { $\rm \delta$ (Reptile)} \\
% \hline
% { 1\%}     & { 29.38} & { 38.69}   & {\textbf{40.06}}   & { + 10.68}        & { + 1.37}            \\
% { 5\%}     & { 33.50}  & { 40.65}   & {\textbf{44.66}}   & { + 11.16}        & { + 4.01}            \\
% { 10\%}    & { 40.94} & { 45.45}   & {\textbf{49.56}}   & { + 8.62}         & { + 4.11}            \\
% { 20\%}    & { 48.09} & { 49.66}   & {\textbf{52.60}}   & { + 4.51}         & { + 2.94}            \\
% { 30\%}    & { 49.27} & { 51.91}   & {\textbf{53.97}}   & { + 4.70}         & { + 2.06}            \\
% { 40\%}       &{ 53.48} & { 53.38}    & {\textbf{56.61}}    & { + 3.13}   & { + 3.23}              \\
% \hline
% \end{tabular}

\begin{tabular}{c|cccccc}
\hline
Percentage & Target Fine-tuning & Reptile    & Task Comb. & Temp. Reptile & { Meta-RTL}  & {  $\rm \delta$ (Target Fine-tuning)} \\ \hline
1\%        & 29.38              & { 38.69} & 38.98            & 38.10      & \textbf{40.06}  & { + 10.68}         \\
5\%        & 33.50              & { 40.65} & 41.72            & 40.35     &\textbf{ 44.66}  & { + 11.16}      \\
10\%       & 40.94              & { 45.45} & 48.78            & 46.82     & \textbf{49.56}     & { + 8.62}       \\
20\%       & 48.09              & { 49.66} & 51.03            & 50.73     & \textbf{52.60}    & { + 4.51}       \\
30\%       & 49.27              & { 51.91} & 50.64            & 52.20      & \textbf{53.97}     & { + 4.70}       \\
40\%       & 53.48              & { 53.38} & 54.26            & 53.38     & \textbf{56.61}      & { + 3.13}     \\ \hline
\end{tabular}

\caption{Accuracy results and improvements on the extremely low-resource settings  on RiddleSense.}
\label{table_small_percent}
\end{table*}

In addition to pure transfer learning, we compared our method with different weight estimation strategies. Both Random and Greedy are based on Reptile. The former randomly generates weights for source tasks while the latter greedily determines weights of source tasks according to rewards, without taking long-term dependency into account (i.e., calculated as $\text{topK} (\bm{r})$, using  rewards from Eq. (\ref{get_reward})). The Random method is worse than Reptile under the unsupervised setting and marginally better than Reptile under the supervised setting but still much worse than Meta-RTL, suggesting that reward signals are important for weight estimation. The Greedy method, in spite of being slightly better than Reptile, is substantially worse than our weight estimation approach in both supervised and unsupervised settings, demonstrating that capturing long-term  dependencies is effective.

\subsection{Evaluation on Extremely Low-Resource Commonsense Reasoning} 

We carried out experiments to evaluate Meta-RTL on extremely low-resource settings. We  randomly selected 1\%, 5\%, 10\%, 20\%, 30\%, 40\% instances from the RiddleSense dataset and used them to form new target datasets. 

Results with BERT are shown in Table \ref{table_small_percent}. First, the smaller the new target dataset is, the larger the improvement of Meta-RTL over the target fine-tuning is gained, demonstrating the capability of the proposed method on extremely low-resource settings. Second, Meta-RTL is better than all three strong baselines on all low-resource settings. Third, Meta-RTL trained with 40\% data of RiddleSense is even better than the target fine-tuning with the entire data by 0.5 points (56.61 vs 56.22).

\subsection{Comparison to Previous Method on Source Task Selection}
% \paragraph{Source Selection} For Meta-RTL, We set up this method to  assign weights during the meta training  process to dynamically adjust the meta training  tasks. Other method use heuristic-based methods to determine how to construct meta training  tasks before meta training , such as Transfer accuracy in \cite{yan-etal-2020-multi-source}. His method can also be considered as a way to modify the weights of the meta training  tasks, but using a prerequisite step to control the selection of source datasets, and during the meta training  process, the meta training  tasks remain treated as equal important.
Previous approaches to multi-source meta-transfer learning usually use a heuristic strategy to select source tasks, e.g., according to the transferability from source tasks to the target task \cite{yan-etal-2020-multi-source}. These methods normally require a preprocessing step to detect suitable source tasks and treat all chosen source tasks equally during meta learning, not allowing
to dynamically adjust the weights of source tasks for meta learning.
\begin{table}[!tbp]
\setlength{\belowcaptionskip}{-12pt} 

\small
\begin{tabular}{c|ccc}
\hline
              & Com2sense & Creak  & RiddleSense \\
\hline
CommonseseQA & - 1.66    & - 0.10 & + 3.82      \\
Winogrande    & + 4.43    & + 0.73 & - 4.71      \\
Com2sense     & 0         & - 0.93 & - 6.44      \\
Creak         & + 5.18    & 0      & - 1.05      \\
RiddleSense   & + 0.70    & - 0.83 & 0  \\
\hline
\end{tabular}
\caption{Transferability results with BERT. The first row displays the target datasets while the first column the source datasets. }
\label{table_transfer}
\end{table}
We compared our method against this static source task selection strategy. First, we  obtained the transferability results for the three  datasets. The results are shown in Table \ref{table_transfer}, where each value denotes the performance change of using the corresponding  dataset in the first column  as the  dataset for pretraining the backbone model and then fine-tuning the pretrained backbone model  on the corresponding target dataset in the first row vs. directly fine-tuning the backbone model on the corresponding target dataset. We compared our method with the transferability-based method using Creak as the target dataset.  
%节省空间Comparison 
Results are shown in Figure \ref{figure_transfer}. For the transferability-based method, we used different combinations of source tasks according to the order of transferability and then ran the meta-transfer learning algorithm described in Section \ref{sec3} where all selected source tasks were treated equally.  

Our model substantially outperforms the transferability-based method under both unsupervised and supervised setting. Our model is better than the best combination by 3.14 points under the unsupervised setting while 3.28 points under the supervised setting.

\begin{figure}[tbp]
\setlength{\belowcaptionskip}{-18pt} 
\centering
\includegraphics[scale=0.4]{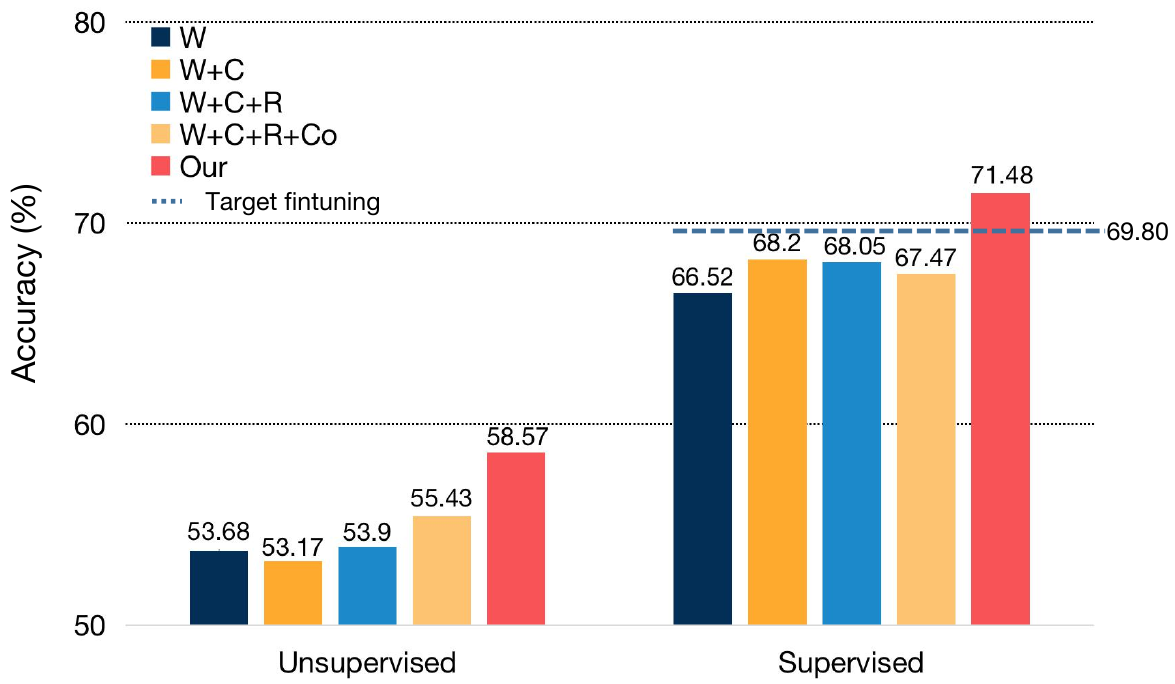}
\caption
{Comparison results of our model vs. the transferability-based method  on Creak. ``C'': CommonseseQA. ``R'': RiddleSense. ``W'': Winogrande. ``Co'': Com2sense.}

\label{figure_transfer}
\end{figure}

\section{Conclusion}
In this paper, we have presented  a reinforcement-based meta-transfer learning framework Meta-RTL  for low-resource  cross-task commonsense reasoning. Meta-RTL uses a reinforcement-based strategy to dynamically estimate the weights of  multiple source tasks for meta and transfer learning from the source tasks to the target task, enabling target-aware weighted knowledge transfer. Our experiments demonstrate the superiority of Meta-RTL over strong baselines and previous static source task selection methods under both unsupervised and supervised settings. Further analyses suggest that Meta-RTL is able to achieve larger improvements over the  target fine-tuning on extremely low-resource settings.

% \section*{Limitations}
% The  proposed framework (Meta-RTL) can be seen as a new and general framework of solving  low-resource tasks by leveraging multiple data sources. which has a wide range of applications not  limited to  commonsense reasoning  discussed in the paper, e.g., machine reading comprehension. However,  this paper only applies and evaluates the proposed framework on low-resource commonsense reasoning. We leave the adaptation of Meta-RTL to other low-resource tasks to our future work. 

% \section*{Ethics Statement}
% Commonsense reasoning has potential positive impact in real-life applications.  While it is expensive and time-consuming to have commonsense reasoning benchmarks for each new task to test the ability of machine, the proposed framework (MetaXCR) can be seen as a new and general paradigm of solving cross-lingual low-resource tasks by leveraging multi-sources, which has a wide range of applications rather than being limited to the commonsense reasoning tasks discussed in the paper, e.g., machine reading comprehension.  From a broader perspective, the proposed framework is not free from some problems of other automatic methods. For example, the model may inherit the biases contained in data. And in our framework, both pre-training and fine-tuning datasets may have biases. Therefore, we encourage future works to study how
% to detect and mitigate similar risks that may arise in our framework.

\bibliography{acl_latex}
\bibliographystyle{acl_natbib}
\appendix
\clearpage

\section{Datasets}
\label{dataset_appendix}

% We conduct experiments to evaluate the performance of MMT on the following Commonsense reasoning benchmark datasets.
\paragraph{CommonseseQA} \cite{talmor-etal-2019-commonsenseqa} is a challenging question answering dataset where  answers are multiple target concepts  that have the same semantic relation to a single source concept from CONCEPTNET. Crowd-sourced workers are asked to author multiple-choice questions that mention the source concept and discriminate in turn between each of the target concepts.
\paragraph{Winogrande} \cite{Sakaguchi_LeBras_Bhagavatula_Choi_2020} is a new dataset with 44K questions, which is  inspired by the original design of WSC, but modified by an  algorithm AFLITE. The algorithm  generalizes human-detectable biases with word occurrences to machine-detectable biases with embedding occurrences to improve the hardness of questions.

\paragraph{Com2sense} \cite{singh-etal-2021-com2sense} is a benchmark dataset which contains 4K complementary true/false sentence pairs. Each pair is constructed with minor perturbations to a sentence to derive its complement such that the corresponding label is inverted.

\paragraph{Creak} \cite{Onoe-Et-Al-2021-Creak} is a testbed for commonsense reasoning about entity knowledge, bridging fact-checking about entities
(e.g., ``Harry Potter is a wizard and is skilled at riding a broomstick.") with commonsense
inferences (e.g., ``if you’re good at a skill you can teach others how to do it.").
\paragraph{RiddleSense} \cite{lin-etal-2021-riddlesense} is a multiple-choice QA dataset which focuses on  the task of answering riddle-style commonsense questions requiring creativity, counterfactual thinking and complex commonsense reasoning.

\begin{table}[tbp]
    \centering
    \begin{tabular}{l|c|ll}
    \bottomrule
      \textbf{Name} & \textbf{\#CA}& \textbf{\#Train}& \textbf{\#Dev}  \\
        \hline 
         CommonseseQA&\centering 5&9,741&1,221\\
         Winogrande&\centering 2&40,398&1,276\\
          Com2sense&\centering 2&1,608&782\\
            Creak&\centering 2&10,176&1,371\\
        RiddleSense&\centering 5&3,510&1,021\\
    \toprule
    \end{tabular}
    \caption{Statistics of the  five  datasets used in our experiments.  CA: candidate answer choices.}
    \label{tab:my_label}
\end{table}
\begin{table}[htb]
    \centering
    \small
    \begin{tabular}{c|cc}
    \bottomrule
        Method&Unsupervised & Fine-tuning \\
        \hline 
        Reptile (BERT) &554/545/549&\multirow{3}{*}{240/245/175}\\
        Temp. Reptile (BERT)&535/515/507&\\
        % Meta-RTL (BERT)&596/595/545&\\
        % Temp. Reptile (BERT) & 535 & 515 & 507& \\
        Meta-RTL (BERT) & 632/633/722 \\
    \toprule
    \end{tabular}
    \caption{500-step runtime (seconds)  of different models on each training stage. The three numbers separated by slash refer to the time consumption of Com2sense / Creak / Riddlesense, respectively.}
    \label{time}
\end{table}
\section{Experimental Setting}
\label{setting_appendix}
We used the BERT-base \cite{devlin-etal-2019-bert} and ALBERT-xxlarge \cite{DBLP:conf/iclr/LanCGGSS20}
as our commonsense reasoning backbone  model. We set the max sequence length to 128. For both  meta training and transfer learning, we adopted  the AdamW optimizer \cite{DBLP:conf/iclr/LoshchilovH19} for Transformers.\footnote{\url{http://github.com/huggingface/transformers}} For meta learning, we set the inner learning rate and outer learning rate to 1e-3 and 1e-5, respectively. The number of inner training iterations was set to 4 and the support batch size for Reptile algorithm was set to 8 for both BERT and ALBERT. For the reinforcement-based weight estimation  module, we used  the policy network similar to  \cite{xiao2020adversarial}. We set the $\epsilon$-greedy rate to 0.2 that used a linear decay toward 0 after 8K steps. The hyper-parameter $K \in \{2, 3\}$ and the temperature hyper-parameter $\omega \in \{1, 2, 5\}$  We used the self-critic algorithm to generate the baseline  value $\Tilde{r}$. All experimental systems were  implemented on Pytorch.
\section{Complexity Analysis}
\label{complexity_appendix}

% \begin{table}[]
%     \centering
%     \begin{tabular}{|l|c|c|}
%     \bottomrule
%         &Unsupervised & Fine-tuning \\
%         \hline 
%         Reptile &554/545/549&\multirow{3}{*}{240/245/175}\\
%         Temperature-based Reptile&535/490/504&\\
%         Meta-TRL&596/595/545&\\
%     \bottomrule
%     \end{tabular}
    
%     \caption{Runtime(s) per 500 steps of different models on each training stage. The three numbers separated by slash refers to the consumption of Com2sense / Creak / Riddlesense, respectively.}
    
%     \label{time_com}
% \end{table}
In order to investigate the additional computational overhead caused  brought by the Meta-RTL framework, we compared the  number of  parameters and running times of Meta-RTL against those of Temperature-based Reptile and Reptile. We show the time consumption of each stage of all methods in Table \ref{time}. The results are estimated  on the same machine by running different methods for 500 steps. It can be seen  that our method requires a slight extra overhead in terms of training time. The only exception is the Riddlesense dataset, which takes longer because it has 5 options and more details can be seen in \ref{riddlesense_reason}.
Additionally, due to the use of parallel computing  in practice, the training time does not increase linearly with the number of datasets used. Hence, our method has  a  high scalability. Regarding additional parameters, all additional parameters are
from the LSTM network. The amount of additional parameters (0.07M) is negligible compared with the number of parameters in BERT/ALBERT. Our method does not affect the convergence of the meta-model.
\end{document}